\pdfoutput=1
\documentclass[runningheads]{llncs}
\usepackage[keeplastbox]{flushend}

\usepackage[ruled,vlined]{algorithm2e}
\usepackage{amsmath,amssymb,amsfonts}
\usepackage{bbding}
\usepackage{booktabs}
\usepackage{cite}
\usepackage{float}
\usepackage{graphicx}
\usepackage{ragged2e}
\usepackage{subfigure}
\usepackage{subfiles}
\usepackage[table,xcdraw]{xcolor}
\usepackage{xtab}

\usepackage{bbding}
\usepackage[marginal]{footmisc}

\begin{document}

\title{CelebHair: A New Large-Scale Dataset for Hairstyle Recommendation based on CelebA}
%
%

\author{Yutao Chen\inst{*} \and
Yuxuan Zhang\inst{*} \and
Zhongrui Huang\inst{} \and
Zhenyao Luo\inst{} \and\\
Jinpeng Chen\inst{(}\Envelope\inst{)}}
\authorrunning{Y. Chen et al.}
%
\institute{School  of  Computer  Science  (National  Pilot  Software Engineering  School) \\
Beijing University of Posts and Telecommunications\\
Beijing, China\\
\email{\{chnyutao, zyuxuan, hzrngu, luozhenyao, jpchen\}@bupt.edu.cn}}

\maketitle              
\footnote{* These authors contributed to the work equllly and should be regarded as co-first authors.\\}

\begin{abstract}
In this paper, we present a new large-scale dataset for hairstyle recommendation, CelebHair, based on the celebrity facial attributes dataset, CelebA. Our dataset inherited the majority of facial images along with some beauty-related facial attributes from CelebA. Additionally, we employed facial landmark detection techniques to extract extra features such as nose length and pupillary distance, and deep convolutional neural networks for face shape and hairstyle classification. Empirical comparison has demonstrated the superiority of our dataset to other existing hairstyle-related datasets regarding variety, veracity, and volume. Analysis and experiments have been conducted on the dataset in order to evaluate its robustness and usability.

\end{abstract}
\section{Introduction}
Hairstyles, together with other facial attributes, affect personal appearance significantly. With the appropriate hairstyle, one's beauty can be enhanced. However, most people and barbers have a very vague understanding of choosing one's suitable hairstyles. Therefore, it is necessary to build a hairstyle recommendation system to recommend the most suitable hairstyle for people, according to their face shapes and other facial attributes. However, the lack of relevant large-scale datasets with essential attributes for hairstyle recommendation has restricted the development of such a system. Satisfactory performance can not be attained based on a limited amount of data with incomplete facial attributes.

To address these problems, we present \textit{CelebHair}, a new large-scale dataset for hairstyle recommendation based on CelebA\cite{liu2015faceattributes}. It consists of 202,599 facial images with the corresponding hairstyles and a wide variety of facial attributes, such as face shape, nose length, and pupillary distance. Together, these attributes contribute to the exploration of the relationship between hairstyle, face shape, and other facial attributes.

CelebHair directly inherits around 200k facial images from the CelebA dataset, along with some hairstyle-related facial attributes. Apart from those, the attributes of our CelebHair dataset are collected from three sources:
\begin{itemize}

\item For extracting features like nose length and pupillary distance, we use a facial landmark detection algorithm, provided by Dlib\cite{dlib09}, followed by algebraic computations to transform and discretize them;

\item For face shape classification, we utilize a pre-trained network of YOLO v4\cite{bochkovskiy2020yolov4}, which can be fine-tuned and customized into a face shape classifier, using an openly accessible face shape dataset on Kaggle\cite{lama_2020};

\item For hairstyle classification, we construct a deep convolutional neural network\cite{lecun1999object} with a spatial transformer\cite{jaderberg2015spatial} substructure for hairstyle classification, based on the training dataset Hairstyle30k\cite{yin2017learning}.
\end{itemize}

To validate CelebHair's usability, we have built a hairstyle recommendation system with the Random Forests\cite{breiman2001random} algorithm as an example application of our dataset. The Random Forests algorithm tries to establish a mapping function from the dependent features like face shape, nose length, and pupillary distance, etc., to the target feature we want to predict -- hairstyle. We have also shown a hairstyle try-on demo application using a combination of facial landmark detection, image rotation, and projection.

Our major contributions in this paper could be phrased as:
\begin{itemize}
    \item To the extent of our knowledge, CelebHair is the first robust, large-scale dataset containing hairstyles, face shapes, and a wide variety of facial attributes. Other existing datasets suffer from defects like having only one gender or including very few features aside from hairstyle.
    \item We have conducted exhaustive experiments around the dataset we have built, forming an applicable framework from hairstyle recommendation to hairstyle try-on. These experiments, in turn, have demonstrated the robustness and usability of our CelebHair dataset.
    \item We share our experience in dataset construction, from feature design to feature extraction. We also presented our applications of facial landmark detection, convolutional neural networks, and spatial transformer networks in the process of feature extraction.
\end{itemize}

\section{Related Works}
\label{section-2}

Related works in this paper mainly involve the following areas: hairstyle datasets, hairstyle recommendation, and other foundational algorithms.

\subsection{Hairstyle Datasets}

We have found several datasets applicable to hairstyle-related tasks. The three most important of them, CelebA, Beauty e-Expert\cite{liu2014wow}, and Hairstyle30k, are introduced here.

CelebA is a dataset of celebrities' facial images with labeled features, some of which, for instance, gender, age, and attractiveness, are considered helpful for our hairstyle recommendation task. CelebA has an advantage in its volume of approximately 200k images. However, only five facial landmarks: left eye, right eye, nose, left mouth, and right mouth, are available. The insufficiency of facial landmarks hinders the extraction of other in-depth facial attributes such as nose length, lip length, and eye width. Moreover, CelebA has very no hairstyle-related features besides hair's waviness and straightness.

Beauty e-Expert is another dataset specifically designed for beauty-related tasks, containing 1505 female photos with various attributes annotated with different discrete values. The Beauty e-Expert dataset covers face shape, facial landmarks, and some other features for clothing and makeup. Nevertheless, three fatal disadvantages have prevented us from adopting the dataset: the tiny volume, the absence of male figures, and the over-simplified hairstyle labels.

Hairstyle30k is a dataset entirely focused on one single attribute: hairstyle. The author has collected the data using keyword searching on online search engines, resulting in a slightly noisy dataset. The original dataset provides a complete variety of 64 hairstyles, with a simplified ten-kind version released afterward. However, no facial attributes other than the hairstyle are included in the dataset.

Domain expert knowledge has suggested that personal hairstyle choices can significantly depend on one's face shape, along with other less critical facial attributes. In a pursuit to build a hairstyle recommendation system, we came to realize the lack of such datasets:
\begin{enumerate}
    \item that are large in volume;
    \item that have labeled face shapes, and other detailed facial attributes;
    \item that cover a wide variety in terms of race, gender, hairstyles, etc.
\end{enumerate}

As all known datasets stated above are defective in at least one of these three aspects, we believe that a new large-scale hairstyle dataset is urgently needed.

\subsection{Hairstyle Recommendation}

Hansini and Dushyanthi\cite{weerasinghe2020machine} have built an end-to-end machine learning framework for hairstyle recommendation, starting from face detection, facial landmark extraction to hairstyle recommendation. Particularly, the Naive Bayes classification algorithm is adopted for hairstyle recommendation by establishing a probabilistic mapping from dependent features to hairstyles. However, the composition and source of the dataset involved in training seem unclear, and the features of the dataset seem relatively inadequate.

Wisuwat and Kitsuchart have done a series of works\cite{sunhem2016approach} \cite{sunhem2016hairstyle} \cite{pasupa2019hybrid} on the subject of hairstyle recommendation. Their dataset's volume seems quite restricted, and only female figures are available. Their strategy for hairstyle recommendation is two-folded:
\begin{itemize}
    \item First, they train a face shape classifier based on facial landmarks. Various machine learning algorithms have been adopted and compared against each other, including SVM, ANN\cite{sunhem2016approach}, and a hybrid of VGG net\cite{simonyan2014very} and SVM\cite{pasupa2019hybrid};
    \item Then, they resort to a set of handcrafted rules summarized from what seems to be domain expert knowledge and empirical opinions for hairstyle recommendation. The entire recommendation procedure only relies on the face shape\cite{sunhem2016hairstyle}.
\end{itemize}

Beauty e-Expert employs a tree-structured probabilistic graph model for hairstyle recommendation, trained upon their own Beauty e-Expert dataset. The model takes into account the relations between attributes from a probabilistic view.

\subsection{Miscellaneous}

Convolutional neural networks, or CNNs for short, have gained attention from image processing researchers worldwide since the uprise of AlexNet\cite{krizhevsky2012imagenet}. CNN is proven to be a potent class of models to consider in any image-related deep learning tasks. In this paper, we use YOLO v4, an effective, optimized member of CNN for real-time object detection, to classify face shapes.

For hairstyle recognition, we use a CNN with a spatial transformer substructure as a hairstyle classifier.The spatial transformer network contains a localization network, which can be conveniently embedded between any two layers of a convolutional neural network. The spatial transformer network takes the input image, transforms it, and then feeds it forward into the next layer. Such transformations render CNNs immune to image translation, scale, rotation, and more generic warping.

Random Forests is an ensemble learning algorithm in which a multitude of decision trees together constitutes a forest. Each decision tree is trained on a subset sampled from the training dataset using the Bootstrap method, and each node in the decision tree may only consider a random subset of all available features to create diversity among base learners. When using random forests for classification, each decision tree in the forest produces a prediction individually. The base learners' predictions are then aggregated using majority voting or weighted voting to form the forest's classification result.
\section{Data Collection}
\label{section-3}

Previous discussions have shown that existing hairstyle-related datasets suffer from insufficiency in either volume or variety. Hence, we introduce CelebHair, a new large-scale dataset for hairstyle recommendation base on CelebA. CelebHair directly inherits around 200k images from CelebA, as well as some hairstyle-related attributes. Techniques including facial landmark detection and convolutional neural networks are also involved to append new facial attributes, for instance, face shape, nose length, and pupillary distance, to those already available in CelebA initially.

Features of the CelebHair dataset come from four sources:
\begin{enumerate}
    \item Features including \textit{eyebrow curve}, \textit{eyeglasses}, \textit{eye bags}, \textit{cheekbone}, \textit{age}, \textit{chubby}, \textit{gender}, \textit{attractiveness}, and \textit{beard} are directly inherited from the CelebA dataset;
    \item Features including \textit{forehead height}, \textit{eyebrow length}, \textit{eyebrow thickness}, \textit{eye width}, \textit{eye length}, \textit{pupillary distance}, \textit{nose length}, \textit{nose-mouth distance}, \textit{lip length}, \textit{lip thickness}, and \textit{jaw curve} are computed from the detected facial landmarks;
    \item \textit{Face shape}s are classified using YOLO v4;
    \item \textit{Hairstyle}s are classified using a CNN with a spatial transformer network substructure.
\end{enumerate}

\begin{figure}[h]
    \centering
    \includegraphics[width=\linewidth]{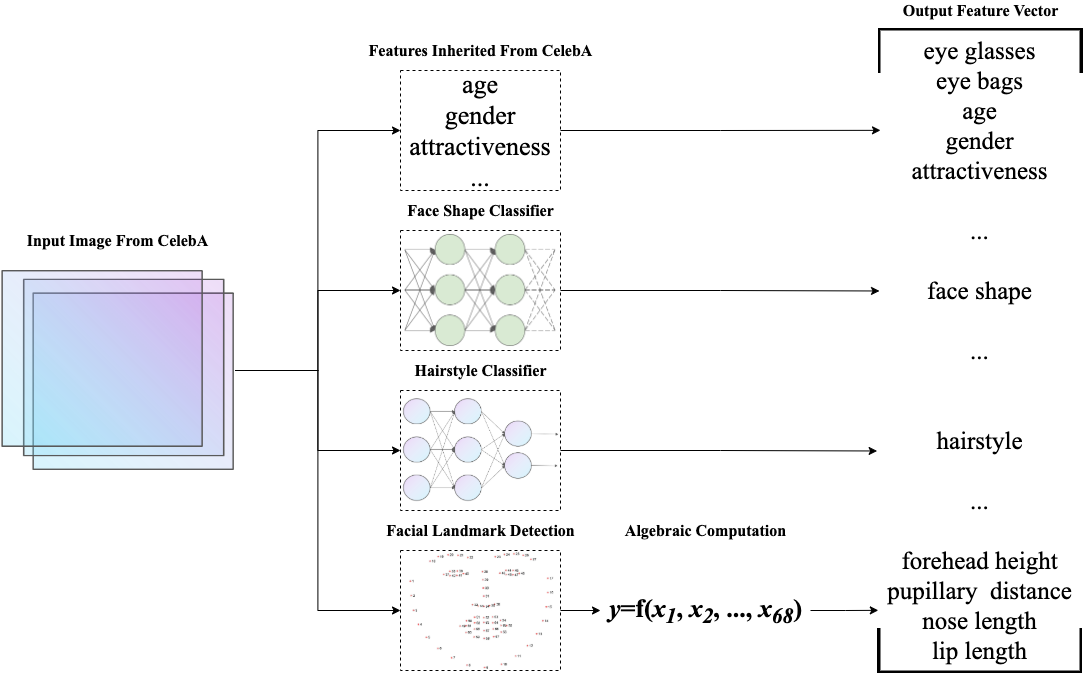}
    \caption{The framework of CelebHair.}
    \label{framework}
\end{figure}

A visual layout of our data collection framework is shown in Fig.~\ref{framework}. Note that the four substeps of data collection could be executed parallelly.

\subsection{Hairstyle Classification}

\begin{figure*}[!ht]
    \centering
    \subfigure[undercut]{\includegraphics[width=2.3cm,height=2.7cm]{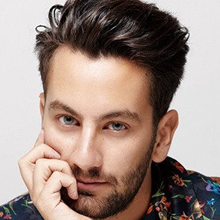}}
    \subfigure[spikyhair]{\includegraphics[width=2.3cm,height=2.7cm]{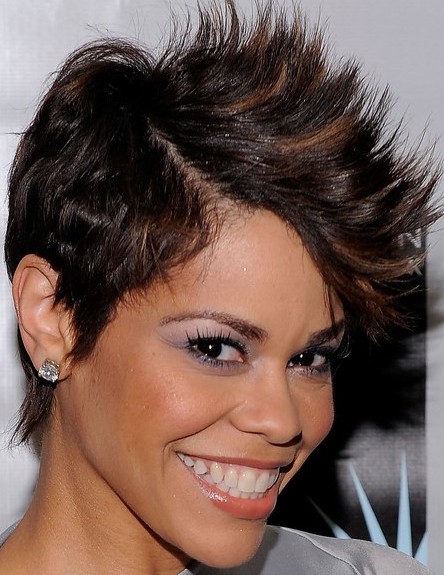}}
    \subfigure[pompadour]{\includegraphics[width=2.3cm,height=2.7cm]{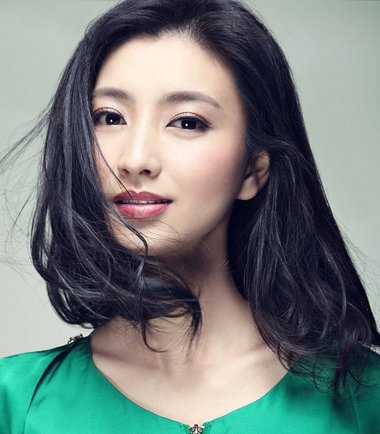}}
    \subfigure[flattop]{\includegraphics[width=2.3cm,height=2.7cm]{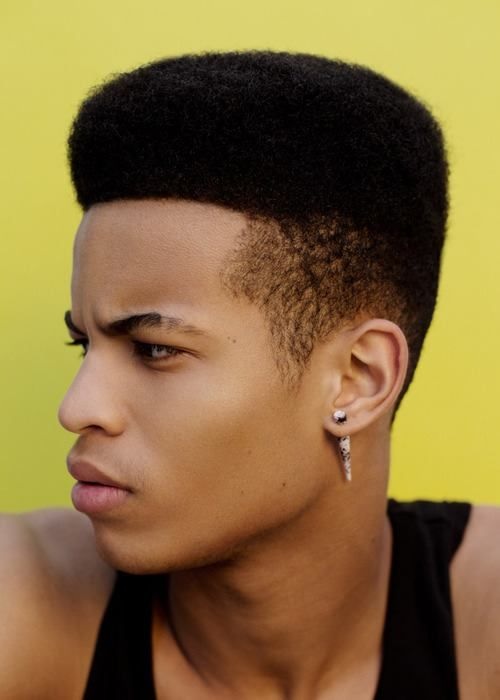}}
    \subfigure[crewout]{\includegraphics[width=2.3cm,height=2.7cm]{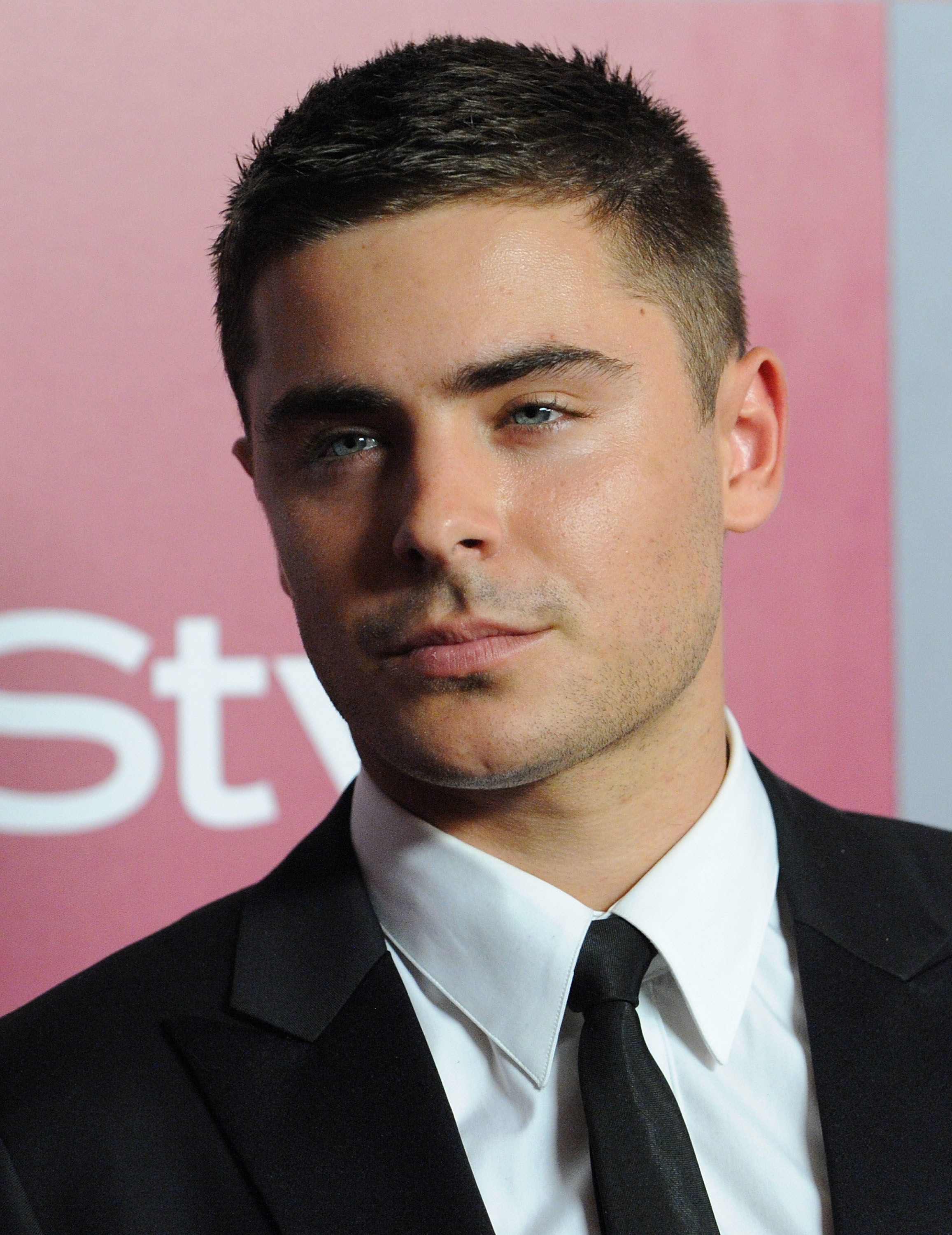}}
    \subfigure[bald]{\includegraphics[width=2.3cm,height=2.7cm]{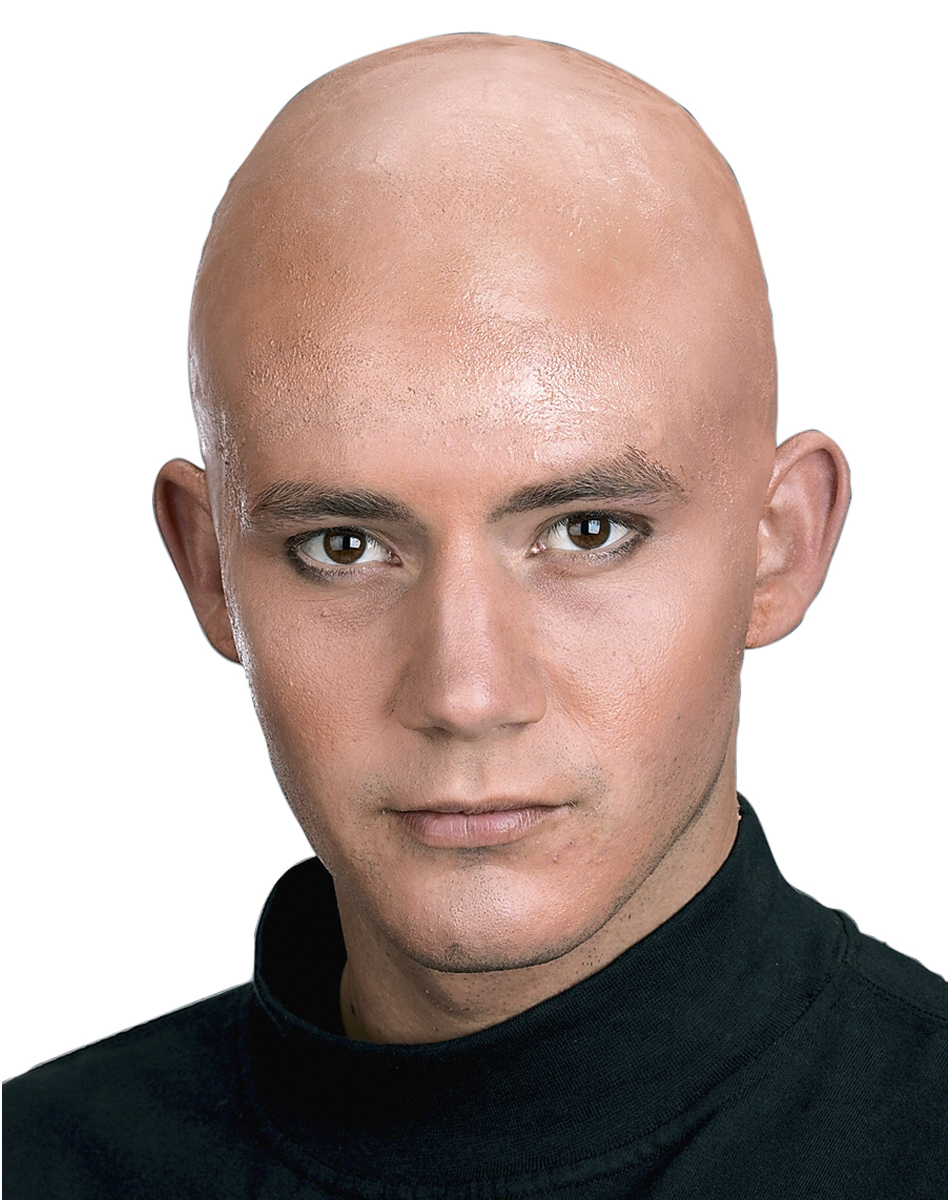}}
    \subfigure[trend curly]{\includegraphics[width=2.3cm,height=2.7cm]{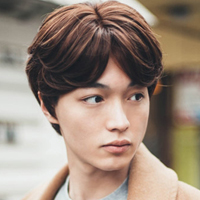}}
    \subfigure[wave]{\includegraphics[width=2.3cm,height=2.7cm]{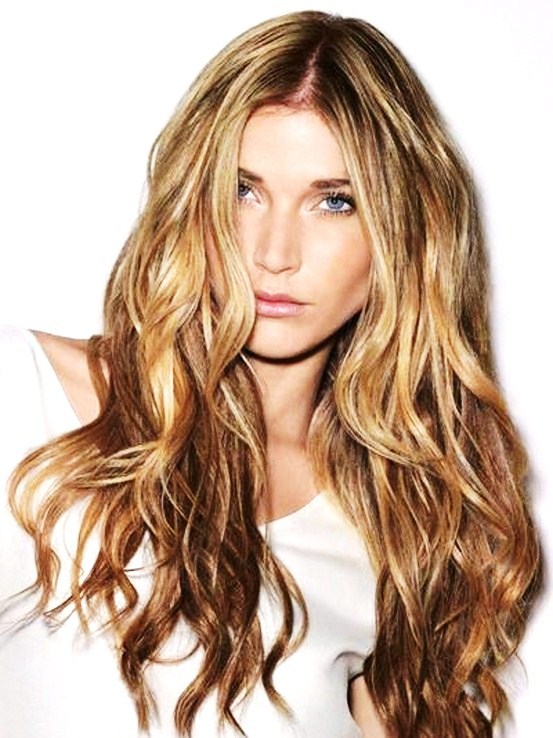}}
    \subfigure[curtained]{\includegraphics[width=2.3cm,height=2.7cm]{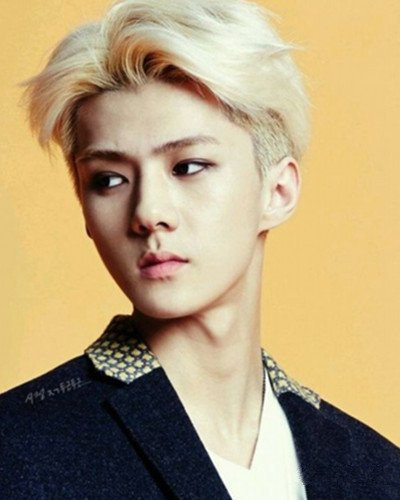}}
    \subfigure[bowlcut]{\includegraphics[width=2.3cm,height=2.7cm]{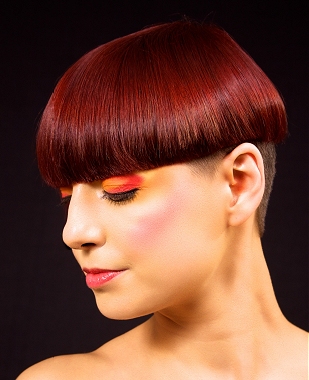}}
    \caption{Ten classes of hairstyles available in the Hairstyle30k dataset.}
    \label{hairstyle}
\end{figure*}

For hairstyle classification, the Hairstyle30k dataset, where ten kinds of most popular hairstyles are available as shown in Fig.~\ref{hairstyle}, is used.

\begin{figure}[H]
    \centering
    \includegraphics[width=\linewidth]{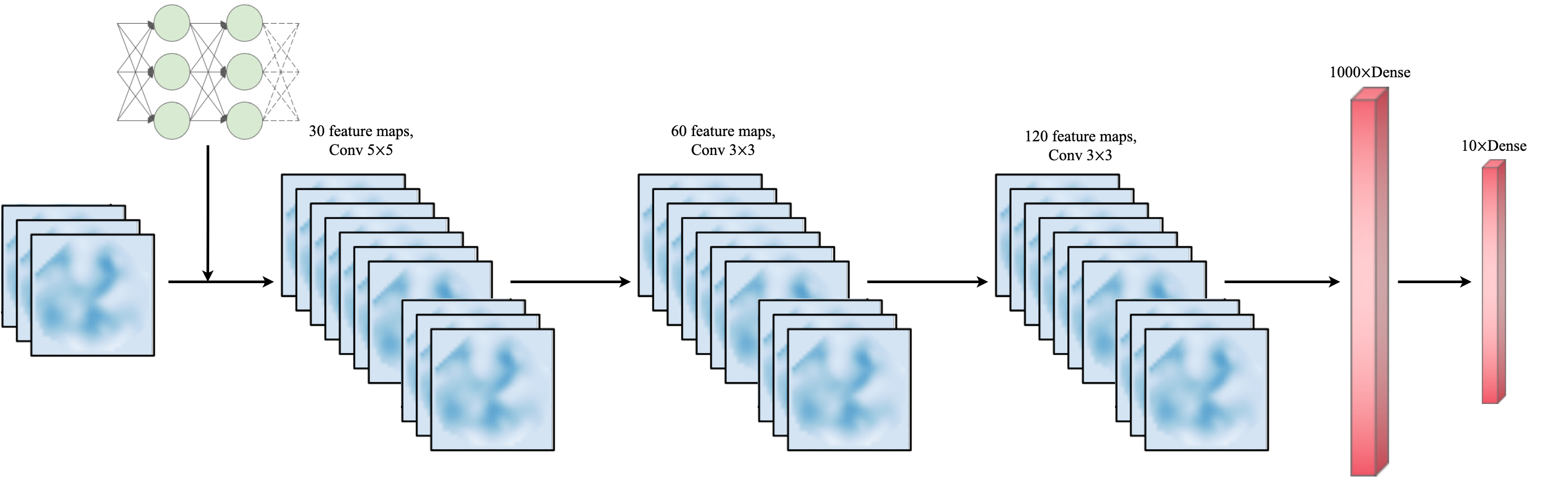}
    \caption{The architecture of our CNN hairstyle classifier.}
    \label{cnn-arch}
\end{figure}

The architecture of the CNN hairstyle classifier is shown in Fig.~\ref{cnn-arch}. The input image comprises three channels (R, G, B) and is feed-forward into three convolutional layers followed by two dense layers. Each convolutional layer has max-pooling and batch normalization\cite{ioffe2015batch}, while all layers, convolutional or dense, are activated by leaky ReLU\cite{maas2013rectifier}. The spatial transformer network substructure is inserted between the input layer and the first convolutional layers, and the localization net of the spatial transformer shares the same three convolutional layers as shown below.

The Hairstyle30k dataset, which contains 10,078 facial images with labeled hairstyles, is randomly shuffled and split into train/test/validation sets at a ratio of 85:10:5. A small validation set is incorporated for early stopping in case of overfitting. With the spatial transformer network (STN), the classification accuracy raises to 85.45\% on the test set. A detailed comparison of the classification performance is given in TABLE~\ref{hairstyle-performance}.

\begin{table}[H]
    \caption{The Accuracy of Hairstyle Classification Models}
    \centering
    \begin{tabular}{c|cc}
        \hline
        \hspace{0em} &\quad Train Acc &\quad Test Acc \\
        \hline
        Without STN &\quad  53.58\% & \quad 42.73\%  \\
        With STN &\quad  88.62\% & \quad 85.45\% \\
        \hline
    \end{tabular}
    \label{hairstyle-performance}
\end{table}

However, the dimension of aligned images provided by Hairstyle30k is $128\times128$, which differs from the dimension of aligned images provided by CelebA, $178\times218$. The mismatch of dimensions has prevented us from feed CelebA facial images straightforwardly into the CNN hairstyle classifier as inputs. As a solution, we crop the CelebA images at the left-upper, left-lower, right-upper, right-lower, and center to retrieve five $128\times128$ images and run the hairstyle classifier on each of them. The results are then aggregated using majority voting to produce a hairstyle class for the corresponding CelebA image.

\subsection{Face Shape Classification}

To extract face shape attributes, we train a face shape classifier, using YOLO v4, an effective real-time object recognition algorithm. It split our image into cells, and each of them predicts several bounding boxes and the probability that there is an object in the bounding box. The face shape dataset we use comprises 5,000 images of female celebrities from all around the globe, categorized according to their face shapes, namely: heart, oblong, oval, round, and square, while each category consists of 1000 images. The training set of each class contains 800 images, whereas the testing set contains 200 images. We mark the bounding box for each of these images, for the original dataset does not include the bounding box of each face.

We train the model based on pre-trained weights with: \textit{batch=64, subdivisions=16, max\_batches = 10,000, steps $\in$ [8000,9000], width=416, height=416, classes=5.}

\begin{figure*}[!ht]
\centering
\subfigure[heart]{
\begin{minipage}[c]{2.3cm}
\centering
\includegraphics[width=2.3cm,height=2.7cm]{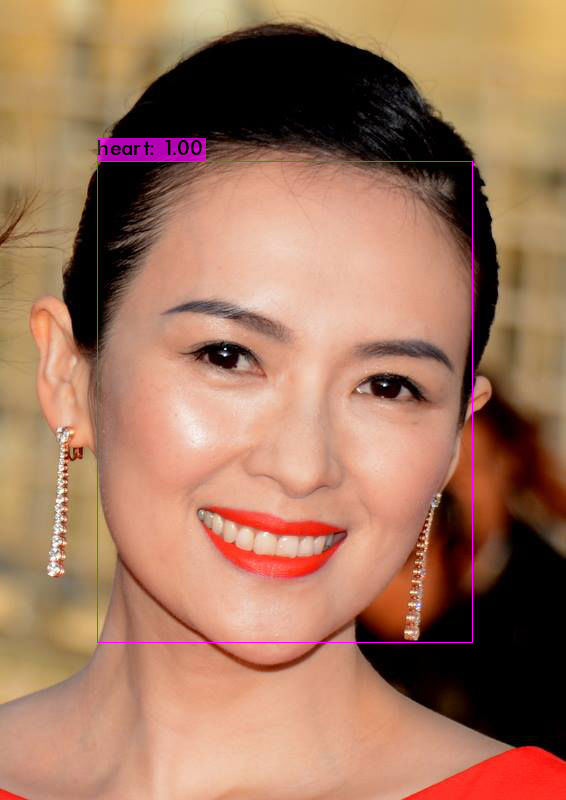}
\includegraphics[width=2.3cm,height=2.7cm]{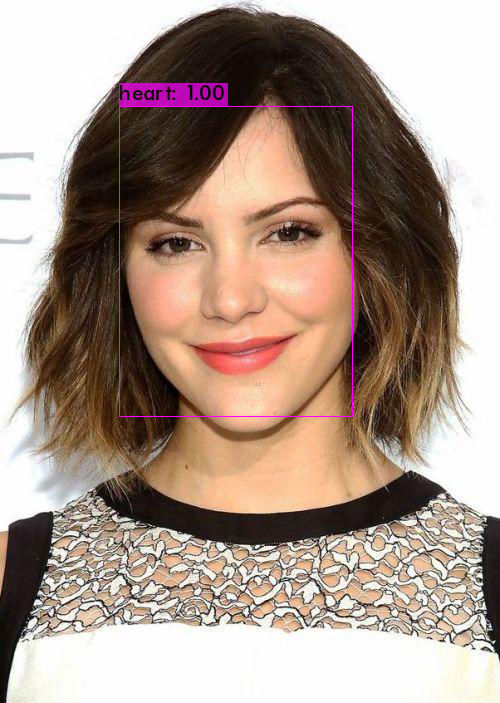}
\end{minipage}%
}%
\subfigure[oblong]{
\begin{minipage}[c]{2.3cm}
\centering
\includegraphics[width=2.3cm,height=2.7cm]{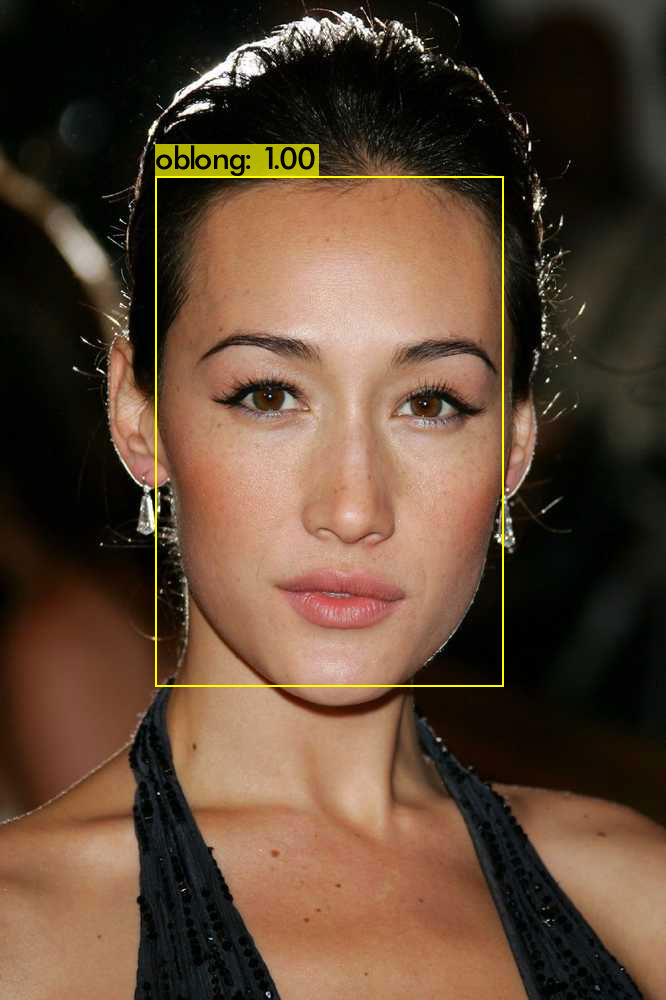}
\includegraphics[width=2.3cm,height=2.7cm]{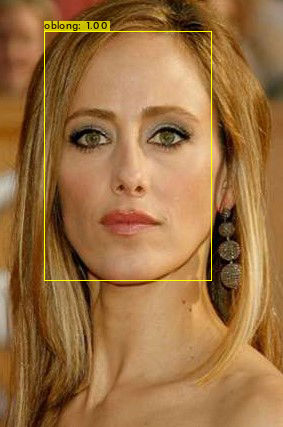}
\end{minipage}%
}%
\subfigure[oval]{
\begin{minipage}[c]{2.3cm}
\centering
\includegraphics[width=2.3cm,height=2.7cm]{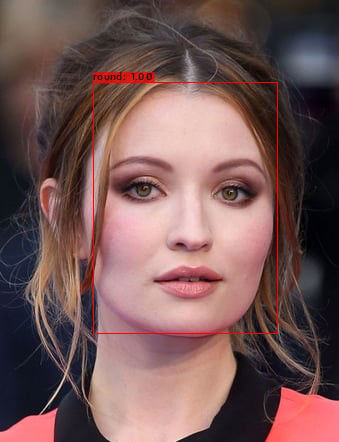}
\includegraphics[width=2.3cm,height=2.7cm]{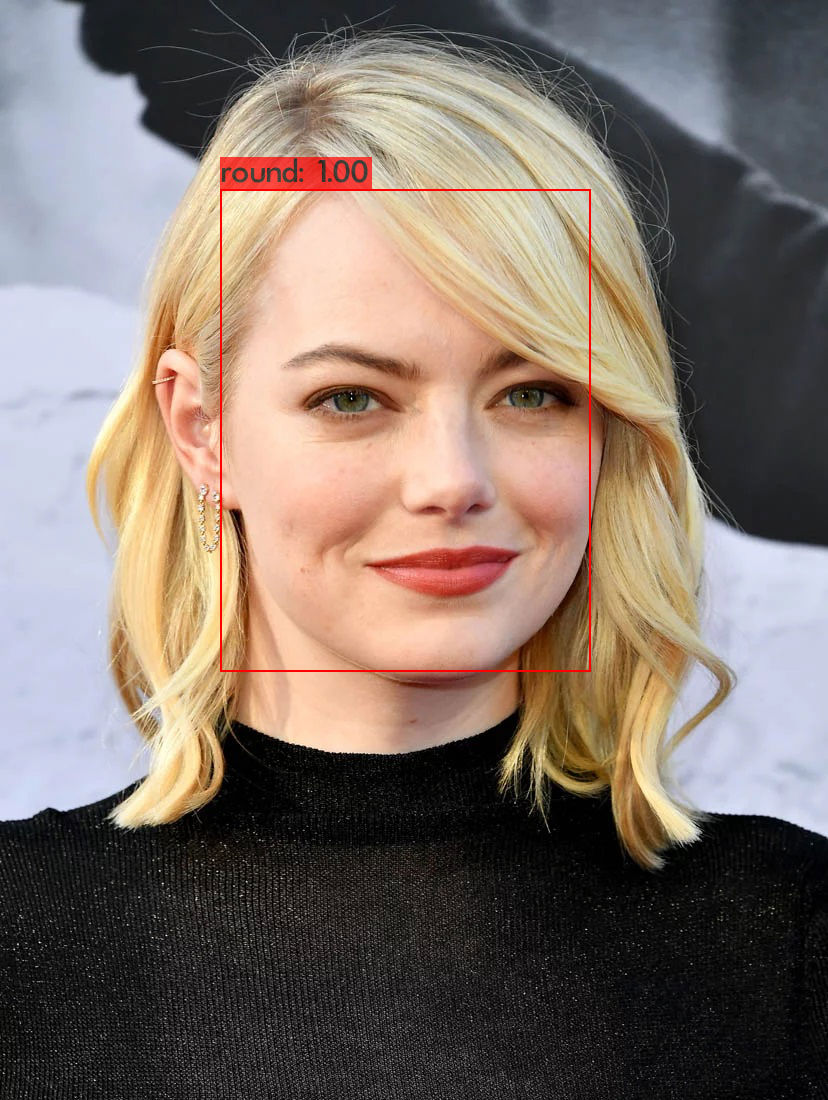}
\end{minipage}
}%
\subfigure[round]{
\begin{minipage}[c]{2.3cm}
\centering
\includegraphics[width=2.3cm,height=2.7cm]{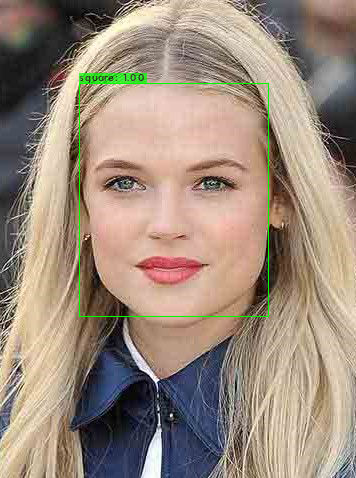}
\includegraphics[width=2.3cm,height=2.7cm]{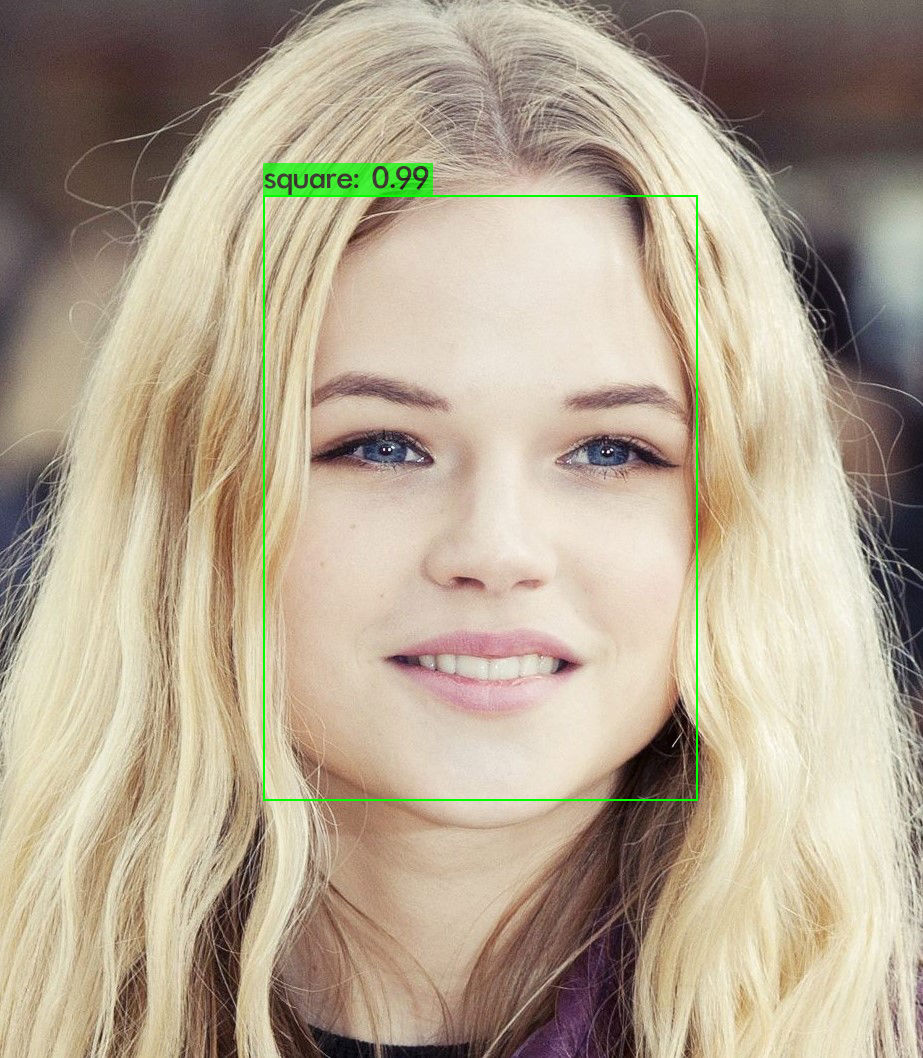}
\end{minipage}
}%
\subfigure[square]{
\begin{minipage}[c]{2.3cm}
\centering
\includegraphics[width=2.3cm,height=2.7cm]{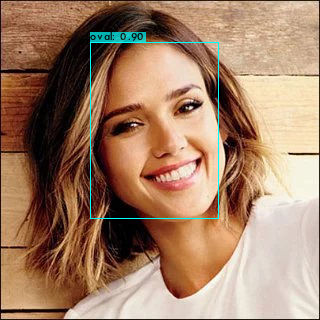}
\includegraphics[width=2.3cm,height=2.7cm]{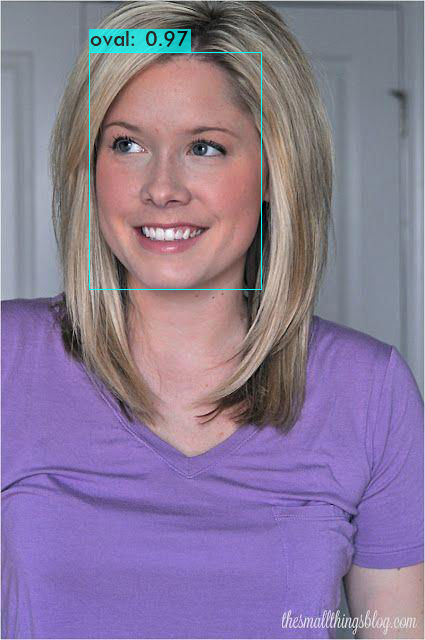}
\end{minipage}
}%
\centering
\caption{Five classes of face shapes and their classification results.}
\label{face-shape}
\end{figure*}

Five classes of face shapes and their recognition results are shown in Fig.~\ref{face-shape}. We evaluate the performance on the face shape classification results' precision as demonstrated in TABLE~\ref{matrix}. Compared to existing work related to face shape classification, we have a much more extensive training size and higher accuracy, as shown in TABLE~\ref{f-com}.

\begin{table}[H]
\centering
\caption{Face Shape Classification Results’ Precision}
\begin{tabular}{c|ccc}
\hline
Label&\quad True Positives&\quad False Positives &\quad Average Precision$^{\mathrm{a}}$\\
\hline
overall & 931& 151&   95.73\%$^{\mathrm{b}}$\\
heart & 186&  17&  98.16\%\\
oblong&  194&  24&  98.46\%\\
oval&  172 & 49 & 90.79\%\\
round&  187 & 41&  93.62\%\\
square & 192&  20&  97.64\%\\
\hline
\multicolumn{4}{l}{$^{\mathrm{a}}$\textit{For confidence threshold = 0.25, false negatives = 69, average }}\\
\multicolumn{4}{l}{\hspace{.5em}\textit{IoU = 79.41\%.}}\\
\multicolumn{4}{l}{$^{\mathrm{b}}$\textit
{IoU threshold = 50\%, used area-under-curve for each unique recall.}}
\end{tabular}
\label{matrix}
\end{table}

\begin{table}[H]
\centering
\caption{Comparison with Other Approaches Towards Face Shape Classification}
\begin{tabular}{c|ccc}
\hline
Approach& Training Set Size& \qquad Accuracy  \\
\hline
SVM-Linear\cite{sunhem2016approach} & 1,000&\qquad 64.00\%\\
SVM-RBF\cite{sunhem2016approach}&  1,000&\qquad 72.00\%\\
MKL with Descriptors\cite{pasupa2019hybrid}& 500&\qquad 70.30\%\\
\textbf{Our Approach} & \textbf{8,000}& \qquad \textbf{87.45\%}\\
\hline
\end{tabular}
\label{f-com}
\end{table}

\subsection{Facial Landmark Detection}

\begin{figure}[]
\centering
\subfigure[]{
\centering
\includegraphics[width=2.7cm,height=3cm]{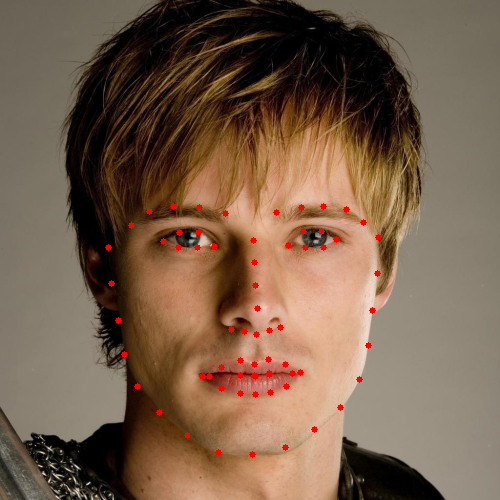}
\label{new-landmark-a}
}%
\subfigure[]{
\centering
\includegraphics[width=2.7cm,height=3cm]{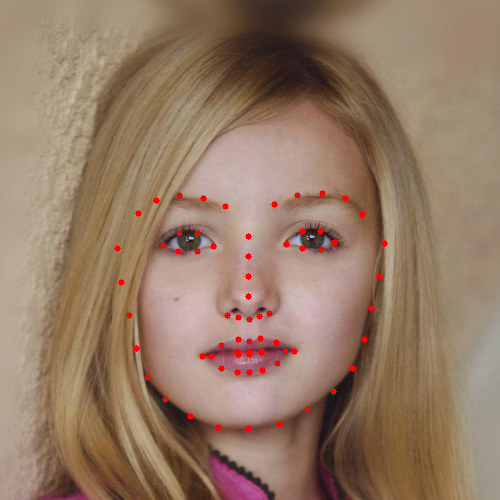}
\label{new-landmark-b}
}%
\subfigure[]{
\centering
\includegraphics[width=2.7cm]{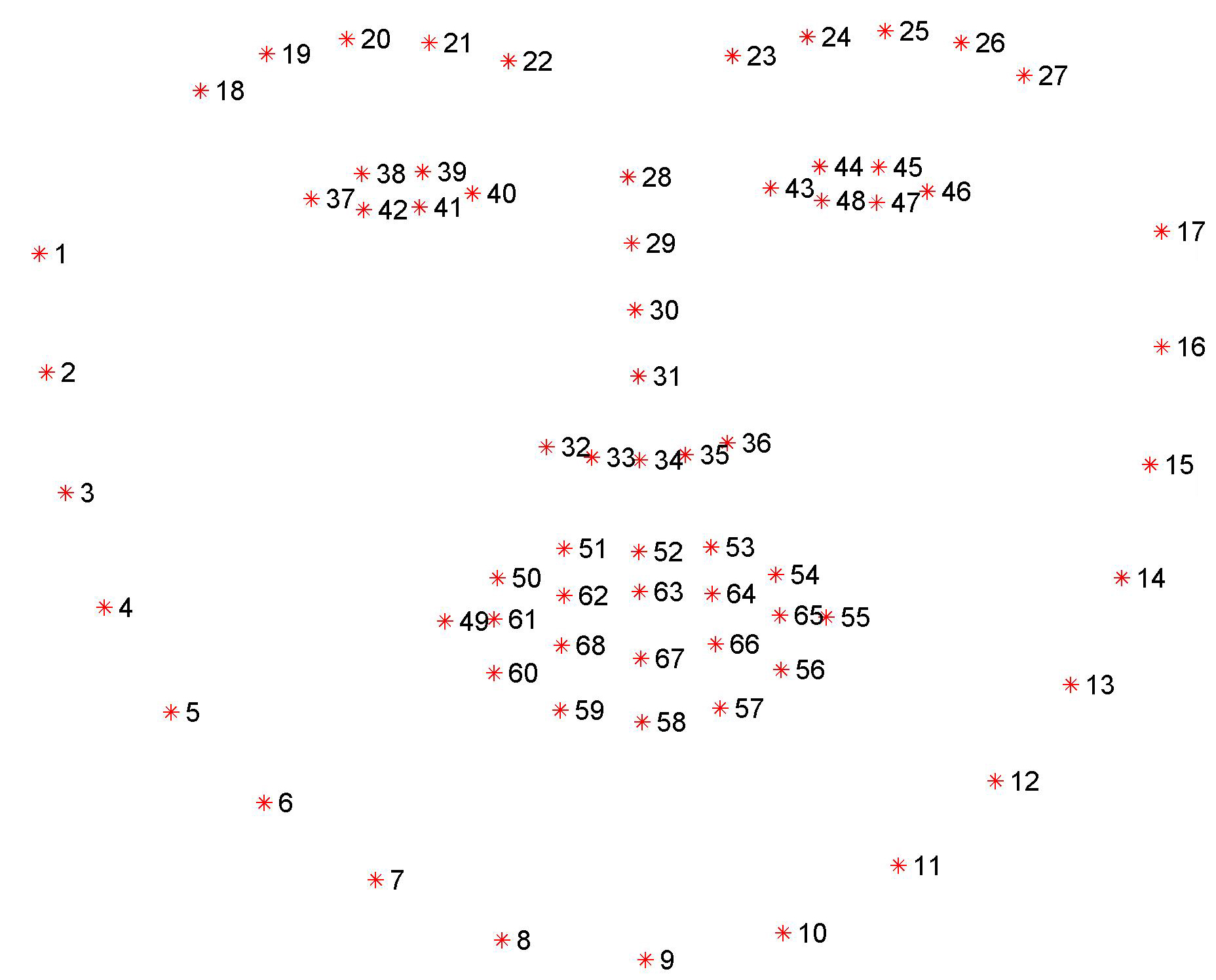}
\label{new-landmark-c}
}%
\centering
\caption{Sixty-eight facial landmark points.}
\label{new-landmark}
\end{figure}

We transform the image into the gray image based on OpenCV\cite{opencv_library}, then extract sixty-eight facial landmarks using Dlib, a toolkit containing machine learning algorithms and tools\cite{dlib09}, as shown in Fig.~\ref{new-landmark-a} and Fig.~\ref{new-landmark-b}. Arithmetic operations can be performed upon these facial landmarks to determine facial attributes, as shown in Fig.~\ref{new-landmark-c}. Here we give a few examples in TABLE~\ref{landmark-calc} on the calculation of such facial attributes, where $d(x,y)$ denotes the distance in pixels between landmark point $x$ and $y$. The attributes, as continuous numerical values, are discretized using equal-width binning at the end.

\begin{table}[]
    \centering
    \caption{Examples for Calculating Facial Attributes Using Facial Landmarks}
    \begin{tabular}{c|c}
        \hline
        Attribute & Formula \\
        \hline
        Forehead Height & $\dfrac{d(18,27)}{d(1,17)}$ \\
        Eye Width & $\dfrac{d(37,40)+d(43,46)}{2\times d(1,17)}$ \\
        Pupillary Distance & $\dfrac{d(42,48)}{d(1,17)}$ \\
        Nose Length & $\dfrac{2\times d(28,34)}{d(22,9)+d(23,9)}$ \\
        Lip Length & $\dfrac{d(51,53)+d(49,55)+d(59,57)}{3\times d(4,14)}$\\
        \hline
    \end{tabular}
    \label{landmark-calc}
\end{table}

\section{Data Description}
\label{section-4}

The CelebHair dataset currently contains 202,599 facial images (inherited from CelebA), and each of these images is labeled with 22 features: forehead height, eyebrow curve, eyebrow length, eyebrow thickness, eye width, eye length, eyeglasses, eye bags, pupillary distance, cheekbone, nose length, nose-mouth distance, lib length, lip thickness, jaw curve, age, chubby, gender, attractiveness, beard, face shape, and hairstyle. A complete list of viable options for each feature is given in TABLE~\ref{attributes}.

\begin{table}[H]
\centering
\caption{Attributes and their variable options}
\begin{tabular}{c|c}
\hline
Attribute & Options\\
\hline
forehead height & short(-1), tall(1)\\
eyebrow curve & straight(-1), curvy(1)\\
eyebrow length & short(-1), long(1)\\
eyebrow thickness & thin(-1), thick(1)\\
eye width & narrow(-1), wide(1)\\
eye length & short(-1), long(1)\\
eyeglasses & none(-1), any(1)\\
eye bags & none(-1), any(1)\\
pupillary distance & short(-1), long(1) \\
cheekbone & low(-1), high(1)\\
nose length & short(-1), long(1)\\
nose-mouth distance & short(-1), long(1)\\
lip length & short(-1), long(1)\\
lip thickness & thin(-1), thick(1)\\
jaw curve & straight(-1), curvy(1)\\
age & young(0), medium(1), old(2)\\
chubby & no(-1), yes(1)\\
gender & female(-1), male(1)\\
attractiveness & no(-1), yes(1)\\
beard & none(-1), any(1)\\
face shape & 1-5, see Fig. \ref{face-shape}\\
hairstyle & 1-10, see Fig. \ref{hairstyle}\\
\hline
\end{tabular}
\label{attributes}
\end{table}

\begin{figure}[H]
    \centering
    \subfigure[Face shape]{
        \includegraphics[width=0.435\textwidth]{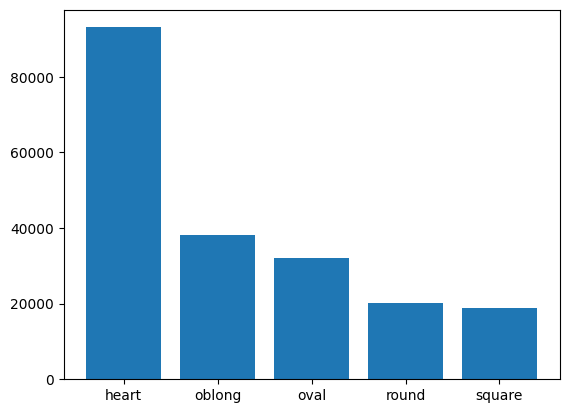}
        \label{face-shape-dist}
    }
    \subfigure[Hairstyle]{
        \includegraphics[width=0.5\textwidth]{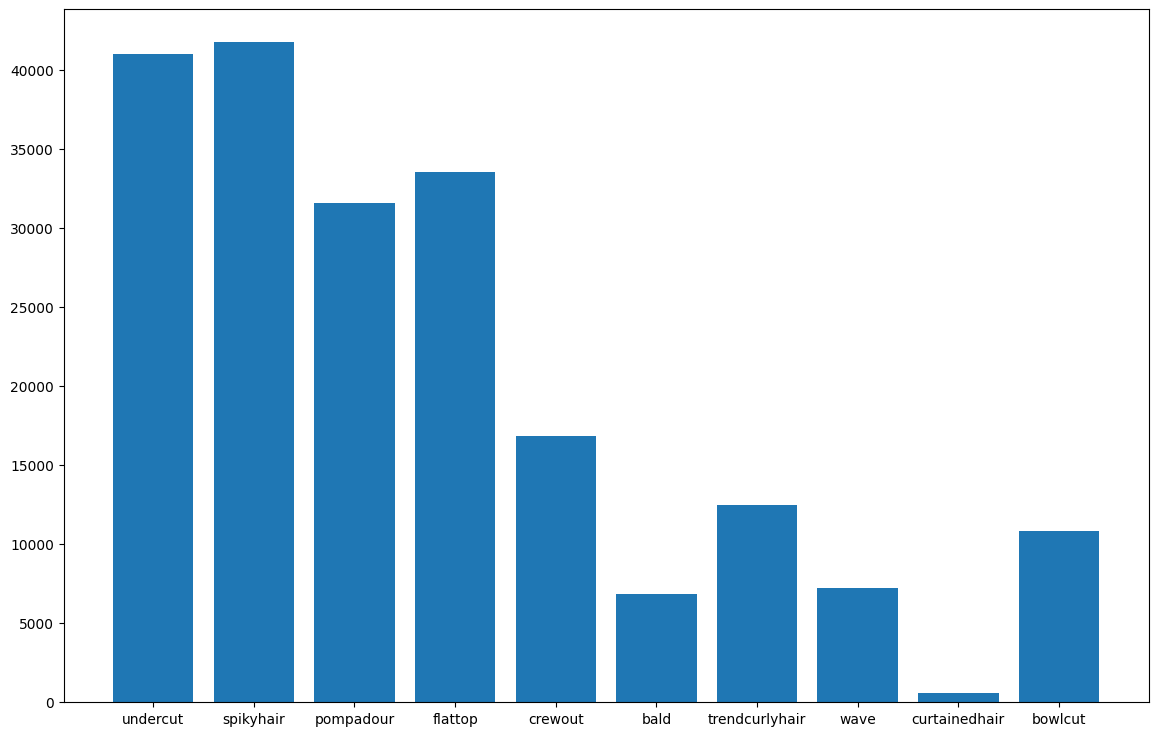}
        \label{hairstyle-dist}
    }
    \caption{The distribution of different face shapes \& hairstyles in \textit{CelebHair}.}
\end{figure}

In Fig.~\ref{face-shape-dist} and Fig.~\ref{hairstyle-dist}, we show the distribution of face shapes and hairstyles within the CelebHair dataset. We see that the majority of celebrity images in CelebA have a heart shape face. Also, undercut and spiky hair seem more popular, while bald and curtained hair seems less so.

We analyze various aspects of CelebHair to provide a deeper understanding of the dataset itself. Comparison with other similar datasets is also available. TABLE~\ref{comparison} shows statistics of CelebHair compared to other datasets, including CelebA, Face Shape Dataset, Hairstyle30k, Beauty e-Expert, AFAD\cite{niu2016ordinal}, Adience\cite{eidinger2014age}, and UTKFace\cite{zhifei2017cvpr}.

\begin{table}[!ht]
\centering
\caption{Comparison between CelebHair and existing similar datasets}
\begin{tabular}{c|c|c|c|c|c|c|c}
\hline
Name&Images&Hairstyle&Face Shape&Featrues&Age&Gender&Landmarks  \\
\hline
CelebA\cite{liu2015faceattributes}&202,599&2&1&9&\Checkmark&\Checkmark&10 \\
Face Shape Dataset\cite{lama_2020}&1000&\XSolid&\Checkmark&\XSolid&\XSolid&\XSolid&\XSolid \\
Hairstyle30k\cite{yin2017learning}&10,078&10&\XSolid&\XSolid&\XSolid&\XSolid&\XSolid\\
Beauty e-Expert\cite{liu2014wow}&1,505&\Checkmark&3&\Checkmark&\XSolid&\XSolid&\XSolid\\
AFAD\cite{niu2016ordinal}&164,432&\XSolid&\XSolid&\XSolid&\Checkmark&\Checkmark&\XSolid \\
Adience\cite{eidinger2014age}&26,580&\XSolid&\XSolid&\XSolid&\Checkmark&\Checkmark&\XSolid \\
UTKFace\cite{zhifei2017cvpr}&23,708&\XSolid&\XSolid&\XSolid&\Checkmark&\Checkmark&68 \\
\textbf{CelebHair}&\textbf{202,599}&\textbf{10}&\textbf{5}&\textbf{17}&\Checkmark&\Checkmark&\textbf{68} \\
\hline
\end{tabular}
\label{comparison}
\end{table}

\section{Application}
\label{section-5}

Example applications of the CelebHair dataset for hairstyle recommendation and hairstyle try-on are present in this section as a demonstration of the dataset's robustness and usability.

\subsection{Hairstyle Recommendation}

For hairstyle recommendation, we have seen the Naive Bayes classification algorithm being used in previous work\cite{weerasinghe2020machine}. Given an adequate dataset with a proper amount of facial attributes and a labeled hairstyle feature, the problem of hairstyle recommendation can be reduced to classification, where a mapping function from the dependent facial attributes (as inputs) to the target hairstyle feature is established. Therefore, we use the Random Forests algorithm for hairstyle recommendation.

The CelebHair Dataset, which carries about 200k records, is shuffled and split into train/test sets at the ratio of 9:1. Then we fine-tune the Random Forests model concerning the number of trees (\textit{n\_estimators}) and the maximum depth of trees (\textit{max\_depth}). Results are shown in TABLE~\ref{random-forest-perf}:

\subsection{Hairstyle Try-on}
There are currently two main methods to implement the integration of hairstyle and face, both of which require hairstyle templates. The first method uses simple image editing, superimposing the hairstyle on the face and manually adjusting the position of the hairstyle\cite{6022427}. As an improvement on the first method, the second method uses a dual linear transformation procedure, in which 21 face contour points are adopted to calculate an affine transformation matrix between the hair template and the input face\cite{liu2014wow}. Therefore, the second method can synthesize the visualized effect of hairstyle automatically. However, both approaches are limited by hair templates, for there are a limited number of hairstyle templates, but there are a variety of hairstyles on different people. 

\begin{table}[H]
    \centering
    \caption{Performance of Random Forests with Different Parameters}
    \begin{tabular}{c|cc}
        \hline
        Hyper Parameters & \quad Train Acc &\quad  Test Acc \\
        \hline
        n\_estimators=5, max\_depth=3 &\quad  37.68\% &\quad  24.37\% \\
        n\_estimators=10, max\_depth=3 &\quad  40.42\% &\quad  29.65\% \\
        n\_estimators=10, max\_depth=5 & \quad 54.39\% &\quad  43.71\% \\
        n\_estimators=20, max\_depth=5 &\quad  77.62\% & \quad 75.43\% \\
        n\_estimators=50, max\_depth=5 &\quad  89.12\% &\quad  87.03\% \\
        n\_estimators=100, max\_depth=5 &\quad  91.29\% &\quad  86.63\% \\
        \hline
    \end{tabular}
    \label{random-forest-perf}
\end{table}

\begin{figure}[H]
\centering
\subfigure[]{
\centering
\includegraphics[width=2.3cm,height=2.7cm]{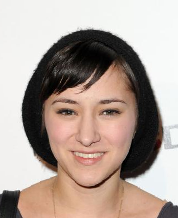}
\label{ft}
}%
\subfigure[]{
\centering
\includegraphics[width=2.3cm,height=2.7cm]{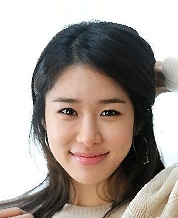}
\label{fi}
}%
\subfigure[]{
\centering
\includegraphics[width=2.3cm,height=2.7cm]{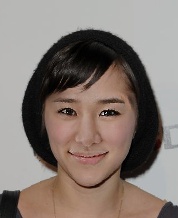}
\label{fo}
}%
\centering
\caption{The hairstyle try-on process.}
\label{face-fusion}
\end{figure}

To address these problems above, we employ a novel approach to visualize the effect of hairstyle try-on and thus intuitively present the outcome of hairstyle recommendation that breaks through the limitations of hairstyle templates. In this approach, we migrate the recommended hairstyles to the input image of the face through face-swapping. We replace facial features on the template image of the face with a recommended hairstyle on it (Fig.~\ref{ft}), with the facial features from the input face image (Fig.~\ref{fi}), and the result (Fig.~\ref{fo}) turns out well.

We implement the face swapping process using an algorithm\cite{matthew} based on Dlib and OpenCV. A formal definition of the algorithm is given in Algorithm~\ref{algorithm1}.

\begin{algorithm}[]
\caption{Hairstyle Try-on Process}
\label{algorithm1}

\SetAlgoLined
\KwIn{Two images, one is the input face image, and the other is the template face image.}
\KwOut{The result of hairstyle try-on.}
\textbf{1:}\quad Extract facial landmarks from the input image and the template image using Dlib, which returns two 68*2 element matrices. Name matrices mentioned above as $i_L$ and $t_L$\;

\textbf{2:}\quad Calculate the centroid from each of the point sets. Name them as $i_C$ and $t_C$\;

\textbf{3:}\quad   $i_L \gets i_L-i_C$\;

\textbf{4:}\quad   $t_L \gets t_L-t_C$\;

\textbf{5:}\quad  Calculate the standard deviation from each of the point sets. Name them as $i_{SD}$ and $t_{SD}$\;

\textbf{6:}\quad  $i_L\gets i_L/i_{SD}$\;

\textbf{7:}\quad  $t_L\gets t_L/t_{SD}$\;

\textbf{8:}\quad  Calculate the rotation portion using the Singular Value. Decompose and return the complete transformation as an affine transformation matrix\;

\textbf{9:}\quad  Change the colouring of the input image to match that of the template image by dividing the input image by a gaussian blur of the input image, and then multiplying by a gaussian blur of the template image\;

\textbf{10:}\quad  Map the input image onto the template image\;

\textbf{11:}\quad  Transfer the facial features from the input face image to the template image\;

\end{algorithm}

In Algorithm~\ref{algorithm1}, firstly, we extract sixty-eight facial landmarks using Dlib, as shown in Fig.~\ref{new-landmark-c}, to locate the part to be extracted from the input face image (Line 1). To make the result turns out well, we align the input image to fit the template image according to the facial landmarks detected (Line 2 - Line 8). Then, we adjust the color balance in the input image to match the template image using OpenCV (Line 9). Finally, we transfer the facial features from the input face image to the template image (Line 10 - Line 11).

\section{Conclusion}
\label{section-6}

In this paper, we have introduced a new large-scale dataset, CelebHair, containing 202,599 celebrity faces with hairstyles, face shapes, and other essential hairstyle-related facial attributes for hairstyle recommendation. Compared with existing similar datasets, CelebHair has a larger volume and more varied features. Consequently, CelebHair provides sufficient support for models that require large-scale data, and should thus refine the performance of hairstyle recommendation models. Moreover, we give several possible applications to illustrate the usability of our dataset, such as building a hairstyle recommendation system with the Random Forests algorithm and visualizing the effect of hairstyle try-on\cite{matthew} using face-swapping. 

Future experiments are planned for further utilizing this dataset. We want to refine the hairstyle try-on experience for users by using Interface GAN\cite{shen2020interpreting}, which enables more natural facial semantic manipulation than simple image rotation and projection.

%
%
%
%
\bibliographystyle{splncs04_unsort}
\bibliography{reference}
\end{document}